\documentclass[]{bytedance_seed}

\usepackage[toc,page,header]{appendix}

\usepackage{minitoc}

\usepackage{amsmath}
\usepackage{amssymb}
\usepackage{graphicx}
\usepackage{booktabs}
\usepackage{tabularx}
\usepackage{algorithm}
\usepackage{algpseudocode}
\usepackage{booktabs}
\usepackage{graphicx}
\usepackage{subcaption}
\usepackage{cuted}
\usepackage{placeins}
\usepackage{booktabs}
\usepackage{tabularx}
\usepackage{graphicx}
\usepackage{caption}
\usepackage{amsmath}
\usepackage{float}

\newcommand{\trendpanel}[2]{%
    \IfFileExists{#1}{%
        \includegraphics[
            width=\linewidth,
            height=0.55\textheight,
            keepaspectratio
        ]{#1}%
    }{%
        \fbox{%
            \parbox[c][0.10\textheight][c]{0.94\linewidth}{%
                \centering\footnotesize #2
            }%
        }%
    }%
}

\title{S\textsuperscript{3}Gym: Can LLMs Turn Self-Testing and Self-Judging into Self-Improvement?}

\affiliation[1]{ByteDance Seed}
\affiliation[2]{M-A-P}
\affiliation[3]{TokenWave.AI}

\contribution{Full author list in Contributions}

\abstract{
Large language models (LLMs) increasingly interact with external environments and accumulate substantial behavioral experience, yet existing agent benchmarks largely evaluate them as fixed policies. It therefore remains unclear whether an agent can actively test its behavior, judge the resulting experience, and use that experience to improve future decisions. We introduce \textbf{S\textsuperscript{3}Gym}, an interactive benchmark for evaluating LLM self-improvement through three coupled capabilities: \textbf{Self-Testing}, \textbf{Self-Judging}, and \textbf{Self-Improvement}. S$^3$Gym separates permissive exploration from strict held-out evaluation and instantiates this protocol in seven text-based games with executable environment verifiers. We evaluate three pathways for incorporating interaction experience: direct History ICL, score-conditioned Summary Memory, and parameter Training.

Our experiments reveal that self-improvement is neither automatic nor uniform. Context-level experience improves performance for several model--game pairs, but the most effective pathway depends strongly on the task structure: summaries are beneficial when experience can be compressed into reusable strategic rules, yet often underperform raw history when success depends on precise, state-contingent information. Parameter training produces substantial gains on some tasks, but also exhibits unstable improvement and severe negative transfer on others. These findings show that recognizing successful actions is insufficient; agents must also transform feedback into executable and transferable policies. S$^3$Gym provides a unified framework for diagnosing this process and identifying the bottlenecks that prevent agents from translating interaction experience into reliable self-improvement.
}

\date{\today}

\checkdata[Project Page]{\url{https://self-developing-agents.github.io/}}

\begin{document}
\maketitle

%不需要目录就注释掉 注意目录不要和第一页放在一块 要有\newpage
%\newpage
%\tableofcontents
%\newpage

\section{Introduction}
\label{sec:introduction}

Large language models (LLMs) are increasingly deployed as autonomous agents that invoke tools, execute code, and interact with external environments through repeated observation--reasoning--action--feedback loops. Such interaction is central to applications including automated research, software engineering, and open-ended decision-making, where agents continually accumulate behavioral experience. Existing benchmarks evaluate multi-turn decision-making, environment exploration, and task completion~\cite{liu2024agentbench,ma2024agentboard}, but they typically treat the evaluated model as a fixed policy. As a result, they primarily answer a static question---how capable is the model at the time of evaluation?---while offering limited insight into whether the model can use its own past interactions to improve future behavior. We refer to this experience-driven capability as \textbf{Self-Improvement}.

Theories of experiential learning and reflective inquiry suggest that experience becomes useful only when it is actively examined, abstracted, and tested again. Dewey and Kolb emphasize the transformation of experience into knowledge through reflection and reapplication~\cite{dewey1933how,kolb1984experiential}, while Popper characterizes progress as the iterative exposure of conjectures to tests that can reveal error~\cite{popper1959logic}. The same principle applies to LLM agents: collecting trajectories alone does not guarantee learning. An agent must generate informative evidence, interpret the causes of success and failure, and convert those interpretations into improved behavior. We therefore decompose experience-driven learning into three interdependent capabilities: \textbf{Self-Testing}, in which the agent explores strategies and gathers diagnostic evidence; \textbf{Self-Judging}, in which it evaluates actions, outcomes, and their potential reusability; and \textbf{Self-Improvement}, in which the resulting experience alters future decisions~\cite{schon1983reflective,argyris1978organizational}.

Experience that has been tested and judged can be incorporated at different levels of persistence. At the short-term \textbf{in-context learning} level, an agent directly conditions future decisions on previous observations, actions, and scores. At the intermediate \textbf{external-memory} level, long trajectories are compressed into natural-language rules, failure patterns, or reusable skills, as in Reflexion and Voyager~\cite{shinn2023reflexion,wang2023voyager}. At the long-term \textbf{training} level, selected or corrected trajectories are internalized through parameter updates, following the broader direction explored by STaR, ReST, and Re-ReST~\cite{zelikman2022star,gulcehre2023rest,dou2024rerest}. These mechanisms make different trade-offs: raw histories preserve detailed evidence but consume context, summaries provide compact abstractions but depend on judgment quality, and training offers persistent internalization but may also amplify incorrectly judged experience. A unified evaluation protocol is therefore needed to compare these pathways under the same interaction and testing conditions.

To address this need, we introduce \textbf{S$^3$Gym}, short for \textbf{Self-Testing, Self-Judging, and Self-Improvement Gym}, an interactive benchmark for evaluating whether LLMs can become more capable through their own environmental experience. Each evaluation consists of an \textbf{Exploration Phase} and an \textbf{Evaluation Phase}, reflecting the distinction between searching for new strategies and exploiting acquired knowledge~\cite{march1991exploration}. During exploration, the agent interacts with relatively permissive environment configurations, generating successful, unsuccessful, and partially successful trajectories. Importantly, S$^3$Gym explicitly separates model-generated self-judgments from environment-verifiable outcomes: self-judgments determine how interaction experience is organized and reused, whereas environment scores measure the resulting behavioral performance. The updated agent is then evaluated on stricter and held-out configurations, providing a direct test of whether experience acquired during exploration transfers to new conditions and whether errors in \textbf{Self-Judging} become a bottleneck to subsequent \textbf{Self-Improvement}.

\begin{figure}[t]
    \centering
    \includegraphics[width=0.72\linewidth]{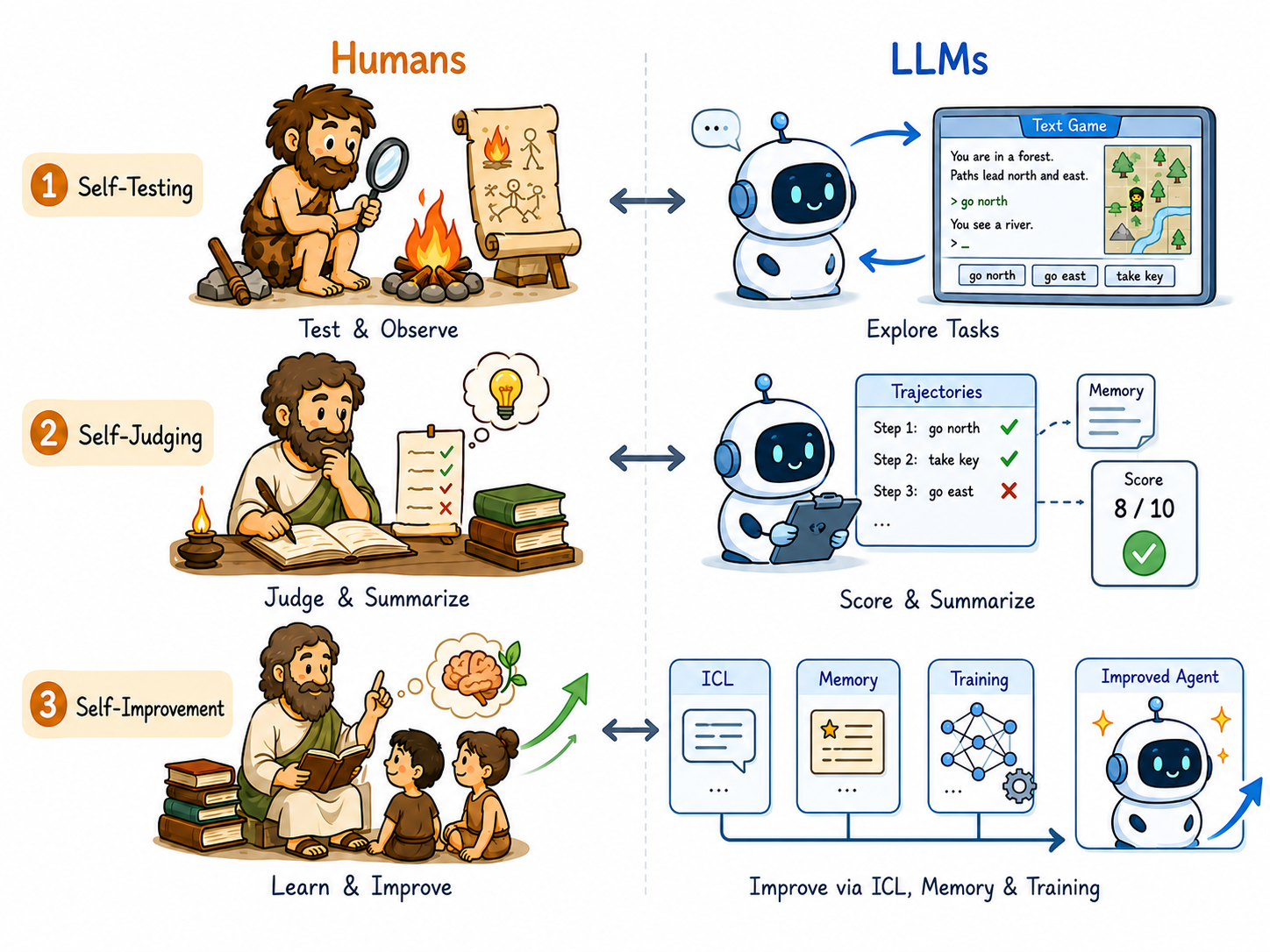}
    \caption{
    Conceptual overview of experience-driven improvement in humans and LLM agents.
    Both processes involve generating experience through Self-Testing, interpreting it through Self-Judging, and incorporating it through reusable memory or learning mechanisms.
    }
    \label{fig:intro_overview}
\end{figure}

We instantiate S$^3$Gym (Figure~\ref{fig:intro_overview}) with text-based games, following the game-based evaluation paradigm of KORGym~\cite{shi2025korgym}. Whereas KORGym evaluates LLM reasoning through diverse interactive games, S$^3$Gym uses such environments as controlled testbeds for \textbf{experience-driven self-improvement}. Their multi-turn interaction, partial observability, delayed consequences, and long-horizon decision-making provide rich behavioral trajectories, while programmatically verifiable rules enable reproducible evaluation with objective step-level or terminal scores. Related environments, including TextWorld, ALFWorld, and ScienceWorld, have similarly demonstrated the utility of textual interaction for evaluating planning, embodied reasoning, and scientific problem solving~\cite{cote2018textworld,shridhar2021alfworld,wang2022scienceworld}. S$^3$Gym includes seven games---\textsc{Chess}, \textsc{Minesweeper}, \textsc{Nullify}, \textsc{Tetris}, \textsc{Snake}, \textsc{PvZ}, and \textsc{Trust Evolution}---covering latent-rule induction, constraint satisfaction, numerical transformation, spatial planning, resource allocation, survival, and multi-agent strategy. Each game provides related but distinct exploration and evaluation configurations, requiring agents to extract transferable knowledge from experience rather than memorize isolated actions or seeds.

\FloatBarrier

Our main contributions are summarized as follows:

\begin{itemize}
    \item \textbf{A unified formulation of experience-driven self-improvement.}
    We formalize LLM self-improvement as an iterative process that connects Self-Testing, Self-Judging, and subsequent behavioral change, and introduce S$^3$Gym with separate exploration and held-out evaluation phases.

    \item \textbf{A systematic comparison of experience-integration pathways.}
    We evaluate three mechanisms for incorporating interaction experience---History ICL, Summary Memory, and parameter Training---under a unified protocol.

    \item \textbf{An explicit evaluation of Self-Judging.}
    By comparing model-generated judgments with verifier-computed outcomes, we assess whether agents can identify useful and harmful experience and translate those judgments into actionable improvement directions.

    \item \textbf{Comprehensive empirical and diagnostic analyses.}
    We evaluate representative LLMs across seven interactive environments and analyze improvement dynamics, judgment reliability, summary effectiveness, cross-game consistency, and common failure modes.
\end{itemize}

\section{Background}
\label{sec:background}

Most agent benchmarks begin after the problem has already been made
executable: the task is specified, the reward or judge is defined, and the
execution scaffold is fixed. This setting is necessary for controlled
comparison, but it hides the work that dominates real deployment. In industry,
that work is distributed across several roles---including solutions
architects, applied engineers, and platform engineers---but its most visible
recent crystallization is the \emph{forward-deployed engineer} (FDE), a title
popularized by Palantir and since adopted by frontier-model
companies~\citep{palantir2020fdse,orosz2025fde}. An FDE is embedded in the
deployment environment and turns a general-purpose model into a system that
works with a customer's data formats, workflows, and operational constraints.
The role's success criterion is not a demonstration or a benchmark score, but
whether the deployed system is genuinely used, continues to work, and improves
in response to failures. The growth of FDE roles across frontier-model and
data-platform companies reflects a simple fact: a capable model is not yet a
working system~\citep{thenewstack2025fde,nanda2025state}, and today the gap is
largely closed by human engineers.

Viewed from the model's side, FDE work supplies three pieces of structure that
benchmark designers normally presuppose. First, the target is vague: an
informal deployment need must be translated into concrete objectives,
constraints, and success criteria. Second, the feedback signal may be absent
or unreliable: tests, judges, traces, or other validation mechanisms must be
constructed before anyone can determine whether the system is improving.
Third, the execution system may not exist in a usable form: the tools, context
management, state, lifecycle logic, and verification interface through which
future tasks will run must be built, adapted, and maintained as requirements
change.

S$^3$Gym focuses on the second layer: the agent-facing feedback signal and the
experience-driven improvement loop that it enables. The benchmark retains an
executable environment verifier as external ground truth, but withholds
verifier-computed outcomes from the agent during exploration. The agent must
instead generate informative evidence through \textbf{Self-Testing}, evaluate
that evidence through \textbf{Self-Judging}, and convert it into better future
behavior through \textbf{Self-Improvement}. This separation reveals both whether
self-judgments agree with actual outcomes and whether judged experience produces
transferable improvement. Whereas \textsc{Aspire} studies how broad deployment
needs become capability growth and \textsc{HarnessDev} studies how models build
and maintain the execution systems that carry them, S$^3$Gym isolates the
experience-to-improvement loop within an executable environment.

\section{Related Work}

\subsection{Interactive Agent Benchmarks}

Recent progress in large language model agents has motivated the development of interactive benchmarks that evaluate multi-turn decision-making, tool use, planning, and environment interaction. AgentBench~\cite{liu2024agentbench} evaluates LLM agents across diverse domains, including operating systems, databases, and games, while AgentBoard~\cite{ma2024agentboard} provides fine-grained analyses of agent progress and failure in multi-step tasks. Text-based environments such as TextWorld~\cite{cote2018textworld}, ALFWorld~\cite{shridhar2021alfworld}, and ScienceWorld~\cite{wang2022scienceworld} further offer controllable testbeds for language-grounded planning, embodied reasoning, and scientific problem solving. Recent game-oriented benchmarks also examine interactive reasoning and strategic behavior in textual or programmatically verifiable environments.

These benchmarks provide rich substrates for evaluating agent behavior, but they generally treat the evaluated model as a fixed policy. Interaction trajectories are primarily used to measure task completion or diagnose failures within a predefined evaluation setting, rather than to determine whether agents can transform their own experience into improved future behavior. S$^3$Gym builds on the controllability and objective verification of interactive environments, while shifting the evaluation target from static task competence to experience-driven behavioral improvement.

\subsection{Experience-Based Agent Improvement}

A growing body of work studies how language agents can improve by reusing their own interaction experience. At the context level, Reflexion~\cite{shinn2023reflexion} introduces verbal reinforcement learning, where agents reflect on previous attempts and use textual feedback to guide future decisions. At the external-memory level, Voyager~\cite{wang2023voyager} accumulates reusable skills in an evolving library, enabling embodied agents to retain and reuse knowledge acquired during open-ended interaction. Recent memory-based approaches similarly compress trajectories into rules, failure patterns, or reusable strategies that can be retrieved in later episodes: ReasoningBank~\cite{ouyang2025reasoningbank} distills generalizable reasoning strategies from an agent's self-judged successful and failed experiences, Tree-of-Experience~\cite{deng2026toe} organizes accumulated experience into a hierarchy that supports feedback attribution and cross-task transfer, and MetaSkill-Evolve~\cite{wang2026metaskill} extends skill-file rewriting to the improvement procedure itself through two-timescale meta-skill evolution.

Experience can also be incorporated through parameter updates. STaR~\cite{zelikman2022star}, ReST~\cite{gulcehre2023rest}, and Re-ReST~\cite{dou2024rerest} improve language models by generating, filtering, or refining self-produced reasoning trajectories. Self-Rewarding Language Models~\cite{yuan2024selfrewarding} further leverage model-generated preference signals for iterative optimization. In interactive environments, RetroAgent~\cite{zhang2026retroagent} trains agents with retrospective dual intrinsic feedback so that experience acquired across episodes can be retrieved and reused rather than only implicitly encoded in parameters, and test-time self-improvement~\cite{acikgoz2025selfimproving} constructs targeted training samples from the agent's own interaction failures. These studies demonstrate that interaction histories and self-generated supervision can improve model performance. However, they typically investigate a specific improvement mechanism within a particular task or training setting, making it difficult to disentangle whether improvement or failure results from the quality of collected experience, the reliability of experience interpretation, or the mechanism used to incorporate that experience. S$^3$Gym provides a unified evaluation protocol that compares different experience incorporation strategies, including History ICL, Summary Memory, and parameter training, under the same exploration and held-out evaluation setting.

\subsection{Evaluating Adaptive and Self-Improving Agents}

The reliability of model-generated supervision has been studied extensively in work on LLM-based evaluation. LLM-as-a-Judge~\cite{zheng2023judging} and JudgeBench~\cite{tan2025judgebench} show that language models can provide useful evaluation signals, but their judgments remain vulnerable to biases, calibration errors, and limitations in reasoning. A recent audit of step-level credit assignment in ALFWorld further shows that LLM-judge scores, outcome-conditioned logprob ratios, and the policy's own confidence all fail to identify causally important steps better than chance when measured against executed replay~\cite{zhang2026credit}. Such limitations are particularly important for self-improving agents, where inaccurate judgments may cause harmful experiences to be retained or valuable experiences to be discarded.

Recent benchmarks have started to move beyond static capability evaluation toward adaptive and self-improving systems. PostTrainBench~\cite{posttrainbench_2026} evaluates automated post-training pipelines that improve model parameters on held-out tasks. SEA-Eval~\cite{sea_eval} studies long-term agent evolution through sequential task interactions, while SEAGym~\cite{seagym} investigates the evolution of persistent agent harnesses, including prompts, memories, tools, and workflows. PAST-Bench~\cite{pastbench_2026} tests whether retained experience actually improves personal agents by running matched task sequences with retained experience turned on and off. ContinualSkillBench~\cite{continualskillbench_2026} measures in-context continual skill learning and finds that gains vary substantially across models and domains, FinEvo-Bench~\cite{finevobench_2026} provides a longitudinal benchmark for self-evolving agents in professional financial workflows, Social Gym~\cite{socialgym_2026} benchmarks multi-agent social reasoning in games with rule-verifiable outcomes, and AI4AI-Bench~\cite{ai4aibench_2026} isolates the design of training algorithms for recursive self-improvement. Beyond benchmarks, recent analyses reveal that self-improvement itself is often unreliable: memory-based methods exhibit high variance and strong sensitivity to task order~\cite{ye2026fragility}, accumulated skills can contaminate later behavior once a defective skill enters the pool~\cite{shang2026backfires}, and optimizer-driven agent gains need not compound across continual-learning rounds~\cite{wang2026compound}. These benchmarks and analyses represent important steps toward evaluating adaptive systems, but they focus on different improvement artifacts or specific aspects of the adaptation process, and none couples self-testing with explicit self-judging under executable environment verification. Table~\ref{tab:comparison} summarizes the key differences between existing benchmarks and S$^3$Gym.

Unlike existing benchmarks that primarily evaluate evolutionary outcomes or individual adaptation mechanisms, S$^3$Gym explicitly studies the process through which agents transform interaction experience into future behavioral improvement. It evaluates multiple experience incorporation strategies within the same environments and interaction budgets, while testing updated agents on disjoint and generally stricter configurations. Moreover, S$^3$Gym records both model-generated judgments and executable environment outcomes, enabling analysis of whether improvement failures originate from inaccurate experience evaluation or from an inability to transform correctly identified feedback into transferable behavior.

\begin{table*}[t]
\raggedright
\small

\caption{Comparison of adaptive and self-improving language agent benchmarks.}
\label{tab:comparison}

\setlength{\tabcolsep}{4pt}
\renewcommand{\arraystretch}{1.15}

\begin{tabularx}{\textwidth}{
    l
    >{\raggedright\arraybackslash}X
    c
    c
    >{\raggedright\arraybackslash}X
    >{\raggedright\arraybackslash}X
}
\toprule
\textbf{Benchmark}
&
\textbf{Updated System}
&
\textbf{Self-Testing}
&
\textbf{Self-Judging}
&
\textbf{Experience Use}
&
\textbf{Evaluation}
\\
\midrule

PostTrainBench
&
Model
&
$\times$
&
$\times$
&
Parameter training
&
Held-out tasks
\\

SEA-Eval
&
Agent
&
Partial
&
$\times$
&
Sequential adaptation
&
Long-term evolution
\\

SEAGym
&
Agent harness
&
Partial
&
$\times$
&
Prompt, memory, tool, workflow updates
&
ID/OOD transfer
\\

PAST-Bench
&
Agent memory
&
Partial
&
$\times$
&
Retained cross-session experience
&
Matched on/off conditions
\\

ContinualSkillBench
&
Skill library
&
Partial
&
$\times$
&
In-context skill accumulation
&
Cross-task reuse
\\

FinEvo-Bench
&
Agent
&
Partial
&
$\times$
&
Longitudinal experience
&
Longitudinal task sequences
\\

Social Gym
&
Agent
&
Partial
&
$\times$
&
Tournament self-play
&
Rule-verifiable outcomes
\\

\midrule

\textbf{S$^3$Gym}
&
Agent behavior
&
$\checkmark$
&
$\checkmark$
&
History ICL, Summary Memory, Training
&
Strict held-out evaluation
\\

\bottomrule
\end{tabularx}
\end{table*}
\FloatBarrier

% Required in the main preamble:
% \usepackage{amsmath,amssymb}
% \usepackage{graphicx}
% \usepackage{booktabs,tabularx}
% \usepackage{algorithm}
% \usepackage{algpseudocode}

\section{Method}
\label{sec:method}

\subsection{Overview}
\label{sec:overview}

S$^3$Gym evaluates whether an LLM can improve its future behavior by testing strategies in interactive environments, judging the resulting experience, and reusing the acquired knowledge. One self-improvement cycle contains five stages:

\begin{equation}
\mathcal{R}_{t}^{(p)} = \mathrm{Explore} \rightarrow \mathrm{Judge} \rightarrow \mathrm{Consolidate} \rightarrow \mathrm{Update}_{p} \rightarrow \mathrm{Evaluate},
\label{eq:overall_cycle}
\end{equation}

where $t$ is the cycle index and $p\in\{\mathrm{History},\mathrm{Memory},\mathrm{Training}\}$ denotes the improvement pathway. During exploration, the agent accumulates trajectories under relatively permissive game configurations. It then judges its own decisions and consolidates the resulting experience into raw history, summary memory, or training examples. Finally, the updated agent is evaluated on stricter and held-out configurations.

\begin{figure}[H]
    \centering
    \IfFileExists{pics/overview.png}{
        \includegraphics[width=0.92\textwidth]{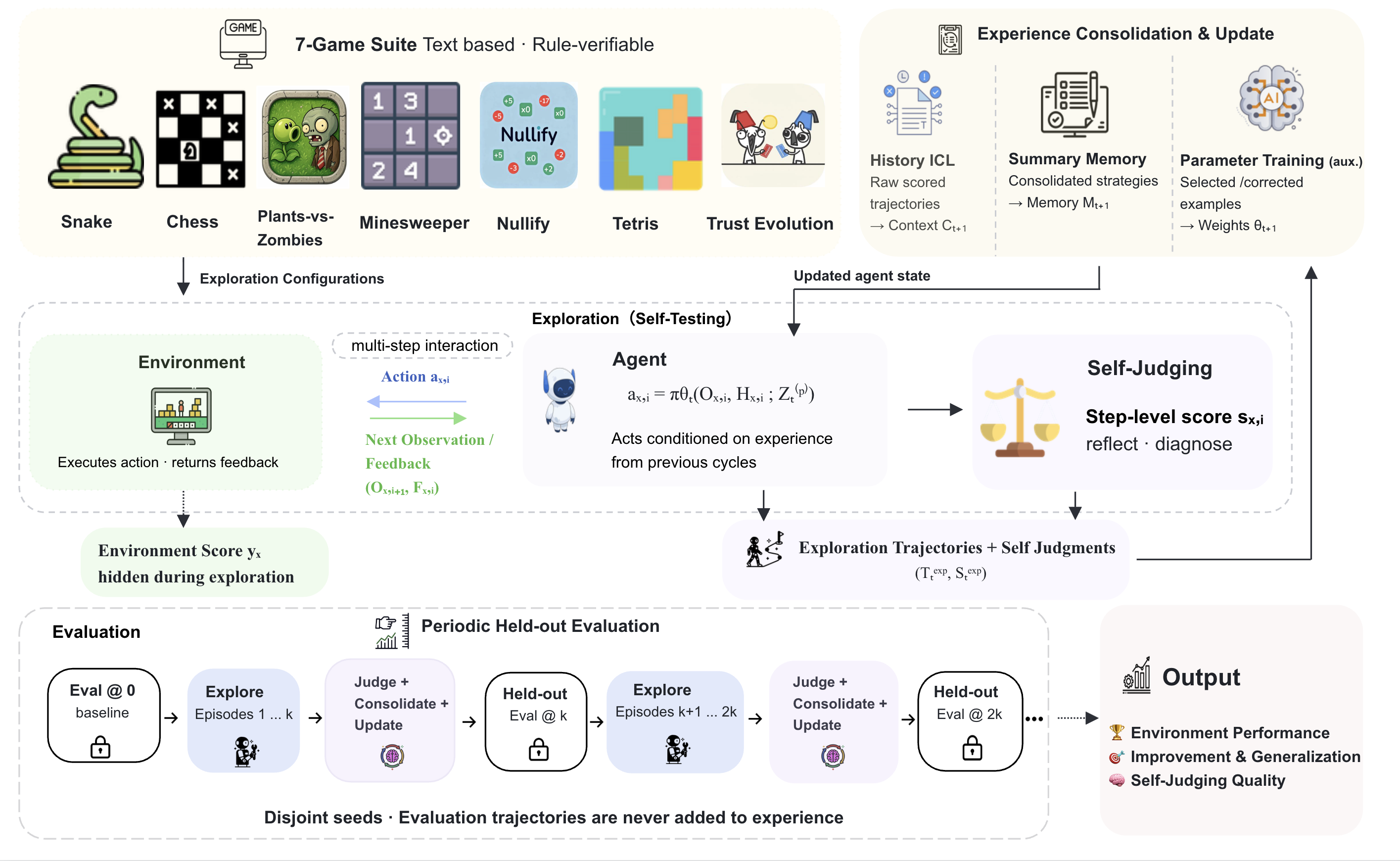}
    }{
        \fbox{\parbox[c][3.2cm][c]{0.92\textwidth}{\centering Method overview figure placeholder}}
    }
    \caption{Overview of S$^3$Gym. Exploration trajectories are judged and incorporated through History ICL, Summary Memory, or parameter Training. The updated agent is then evaluated on stricter game configurations.}
    \label{fig:method_overview}
\end{figure}

\subsection{Interaction Protocol and Formalization}
\label{sec:interaction}

Each game contains multiple independent episodes, and each episode consists of a sequence of interaction steps. For episode \(x\) at step \(i\), we denote the current observation by \(O_{x,i}\), the within-episode interaction history by \(H_{x,i}\), the agent action by \(a_{x,i}\), the agent's self-judged immediate score by \(s_{x,i}\), the observable environment feedback by \(F_{x,i}\), the verifier-computed step reward by \(r_{x,i}\), and the final environment score by \(y_x\).

An episode is initialized as

\begin{equation}
O_{x,0}=\operatorname{Reset}(\operatorname{seed}_x,c_x),
\end{equation}

\begin{equation}
H_{x,0}=\varnothing,
\end{equation}

where \(c_x\) denotes the game configuration.

At each step, the agent receives the current observation, the within-episode history, and the pathway-specific experience state \(Z_t^{(p)}\) inherited from previous exploration episodes. Following the actual interaction interface, the model jointly produces an action and a self-judged immediate score:

\begin{equation}
(a_{x,i},s_{x,i})
=
\pi_{\theta_t}
\left(
O_{x,i},H_{x,i};Z_t^{(p)}
\right),
\end{equation}

where \(p\in\{\text{History},\text{Memory},\text{Training}\}\). The score \(s_{x,i}\) represents the model's estimate of the immediate reward associated with its proposed action under the game rules, rather than a free-form confidence score. In the context-level pathways, \(Z_t^{(p)}\) contains either retained raw history or summarized experience; for parameter training, acquired experience is instead reflected primarily through the updated parameters \(\theta_t\).

The environment then executes the proposed action and returns the next observation, observable feedback, verifier-computed reward, and termination signal:

\begin{equation}
(O_{x,i+1},F_{x,i},r_{x,i},d_{x,i})
=
\operatorname{Env}(O_{x,i},a_{x,i}).
\end{equation}

Importantly, \(F_{x,i}\) denotes environment information available to the agent for subsequent interaction, whereas \(r_{x,i}\) is retained by the benchmark as an external verification signal. In the main setting, verifier-computed rewards are not exposed to the agent during exploration. This separation allows S\(^3\)Gym to compare the agent's self-judgment \(s_{x,i}\) against the actual environment reward \(r_{x,i}\) without allowing ground-truth rewards to directly guide subsequent decisions.

The agent-visible transition is then appended to the within-episode history:

\begin{equation}
H_{x,i+1}
=
H_{x,i}
\oplus
\operatorname{Format}
\left(
O_{x,i},
a_{x,i},
s_{x,i},
F_{x,i},
O_{x,i+1}
\right),
\end{equation}

where \(\oplus\) denotes ordered concatenation. Thus, the interaction history records the agent's own score together with the state transition, while the verifier reward \(r_{x,i}\) remains benchmark-side information. At the next step, the agent again applies Eq. (4) to the updated observation and history.

When the episode terminates or reaches the maximum number of steps, the environment returns the final verifier-computed score

\begin{equation}
y_x
=
\operatorname{Score}(\tau_x),
\end{equation}

where \(\tau_x\) denotes the complete executed trajectory of episode \(x\). The step-level reward \(r_{x,i}\) is used to evaluate the reliability of Self-Judging, whereas \(y_x\) measures the episode-level behavioral performance used for benchmark evaluation. Neither signal is provided to the agent during exploration in the main setting.

% \begin{algorithm}[!htbp]
% \small
% \caption{One S$^3$Gym Episode}
% \label{alg:episode}
% \begin{algorithmic}[1]
% \Require Game configuration $c_x$, seed $\mathrm{seed}_x$, experience state $Z_t^{(p)}$
% \State $O_{x,0} \gets \operatorname{Reset}(\mathrm{seed}_x,c_x)$
% \State $H_{x,0} \gets \varnothing$
% \For{$i=0,\ldots,I_{\max}-1$}
%     \State $a_{x,i} \gets \pi_{\theta_t}(O_{x,i},H_{x,i};Z_t^{(p)})$
%     \State $(O_{x,i+1},F_{x,i},d_{x,i}) \gets \operatorname{Env}(O_{x,i},a_{x,i})$
%     \State $s_{x,i} \gets J_{\theta_t}(O_{x,i},a_{x,i},F_{x,i},O_{x,i+1},H_{x,i};Z_t^{(p)})$
%     \State $H_{x,i+1} \gets H_{x,i} \oplus \operatorname{Format}(O_{x,i},a_{x,i},F_{x,i},O_{x,i+1},s_{x,i})$
%     \If{$d_{x,i}=1$}
%         \State \textbf{break}
%     \EndIf
% \EndFor
% \State $y_x \gets \operatorname{Score}(H_{x,m_x})$
% \State \Return $(H_{x,m_x},y_x)$
% \end{algorithmic}
% \end{algorithm}
% \FloatBarrier

\subsection{Self-Judging and Experience Consolidation}
\label{sec:self_judging}

Self-Judging provides an internal signal for identifying useful and harmful experience when precise environment rewards are unavailable to the agent. Low-scoring trajectory segments indicate possible mistakes or strategies to avoid, whereas high-scoring segments provide candidate strategies to retain.

S$^3$Gym organizes the judged experience in two context-level forms.

\paragraph{History Context.}
Score-annotated trajectories are directly serialized and appended to future contexts:

\begin{equation}
C_{t+1} = \operatorname{Serialize}\!\left(C_t,\mathcal{T}_{t}^{\mathrm{exp}},\mathcal{S}_{t}^{\mathrm{exp}}\right),
\label{eq:history_context}
\end{equation}

where $\mathcal{T}_{t}^{\mathrm{exp}}$ and $\mathcal{S}_{t}^{\mathrm{exp}}$ are the exploration trajectories and their self-judgment scores in cycle $t$.

\paragraph{Summary Memory.}
The model compresses the judged trajectories into reusable experience:

\begin{equation}
M_{t+1} = \operatorname{Summarize}\!\left(M_t,\mathcal{T}_{t}^{\mathrm{exp}},\mathcal{S}_{t}^{\mathrm{exp}}\right).
\label{eq:summary_memory}
\end{equation}

The summary contains strategies to retain, mistakes to avoid, and concrete directions for the next interaction cycle:

\begin{equation}
M_{t+1} = \left(R_{t}^{\mathrm{retain}},R_{t}^{\mathrm{avoid}},D_{t}^{\mathrm{next}}\right).
\label{eq:memory_schema}
\end{equation}

\subsection{Exploration and Evaluation Phases}
\label{sec:exploration_evaluation}

Each game provides an exploration distribution $\mathcal{C}_{g}^{\mathrm{exp}}$ and an evaluation distribution $\mathcal{C}_{g}^{\mathrm{eval}}$. Exploration environments preserve the main task structure but use more permissive conditions, such as smaller state spaces, weaker failure penalties, or additional trial opportunities. Evaluation environments use stricter rules or more complex configurations.

A self-improvement cycle contains $N_{\mathrm{exp}}$ exploration episodes followed by $N_{\mathrm{eval}}$ evaluation episodes:

\begin{equation}
\left\{\tau_{t,x}^{\mathrm{exp}}\right\}_{x=1}^{N_{\mathrm{exp}}} \rightarrow \operatorname{Improve} \rightarrow \left\{\tau_{t,x}^{\mathrm{eval}}\right\}_{x=1}^{N_{\mathrm{eval}}}.
\label{eq:phase_schedule}
\end{equation}

Exploration provides diverse experience for judgment and consolidation, whereas evaluation verifies whether the resulting improvement transfers to stricter conditions. The two phases use disjoint random seeds, and evaluation trajectories are not added to history, memory, or training data.

For pathway $p$ and game $g$, evaluation performance after cycle $t$ is

\begin{equation}
\overline{Y}_{t,g}^{(p)} = \frac{1}{N_{\mathrm{eval}}}\sum_{x=1}^{N_{\mathrm{eval}}}y_{t,g,x}^{(p)},
\label{eq:evaluation_score}
\end{equation}

and improvement over the original model is

\begin{equation}
\Delta_{t,g}^{(p)} = \overline{Y}_{t,g}^{(p)} - \overline{Y}_{0,g}.
\label{eq:improvement}
\end{equation}

We report both the final evaluation score and $\Delta_{t,g}^{(p)}$. The former measures final capability, while the latter measures how effectively the model benefits from its own interaction history.

\subsection{Self-Improving Pathways}
\label{sec:pathways}

S$^3$Gym evaluates three pathways for incorporating exploration experience.

\paragraph{History ICL.}
Score-annotated trajectories are directly inserted into the context of later episodes:

\begin{equation}
Z_{t+1}^{(\mathrm{History})} = C_{t+1}.
\end{equation}

This pathway preserves detailed interaction evidence but is constrained by context length.

\paragraph{Summary Memory.}
The agent receives the consolidated summary rather than the complete trajectory:

\begin{equation}
Z_{t+1}^{(\mathrm{Memory})} = M_{t+1}.
\end{equation}

This pathway evaluates whether the model can abstract transferable strategies from lengthy interaction histories.

\paragraph{Parameter Training.}
Exploration trajectories and self-judgments are converted into supervised fine-tuning examples:

\begin{equation}
\mathcal{D}_{t}^{\mathrm{SFT}} = \operatorname{BuildSFT}\!\left(\mathcal{T}_{t}^{\mathrm{exp}},\mathcal{S}_{t}^{\mathrm{exp}}\right),
\end{equation}

followed by the parameter update

\begin{equation}
\theta_{t+1} = \operatorname{SFT}\!\left(\theta_t,\mathcal{D}_{t}^{\mathrm{SFT}}\right).
\end{equation}

High-scoring actions are retained as positive examples, while low-scoring actions are filtered or replaced with corrected actions. Training is treated as an auxiliary pathway because it introduces optimization variables absent from context-level improvement.

All pathways start from the same base model and use the same exploration seeds, evaluation seeds, interaction budget, and decoding settings.

%\subsection{Comparison with Existing Benchmarks}
%\label{sec:comparison}

%Existing gym-style benchmarks mainly evaluate general reasoning, competitive gameplay, or reinforcement learning with verifiable rewards. S$^3$Gym instead focuses on whether an agent can improve by judging and reusing its own interaction trajectories.

%In summary, S$^3$Gym evaluates an endogenous improvement loop: the agent generates experience through Self-Testing, interprets that experience through Self-Judging, and incorporates it through History ICL, Summary Memory, or parameter Training. Improvement is successful only when it produces higher scores on held-out evaluation episodes.

% Required packages:
% \usepackage{booktabs,tabularx,graphicx,caption}

\section{Experiments}
\label{sec:experiments}

\subsection{Benchmark Games}
\label{sec:games}

S$^3$Gym contains seven text-based games that cover complementary forms of interactive reasoning, including latent-rule induction, constraint satisfaction, numerical transformation, spatial planning, long-horizon control, resource allocation, and multi-agent strategy. Each game provides an exploration configuration and an evaluation configuration. The exploration configuration is generally more permissive, while the evaluation configuration introduces stricter termination conditions, a larger search space, or denser dynamics. Table~\ref{tab:env-setting} summarizes the code-level configurations used in our experiments.

\begin{table}[H]
\centering
\small
\setlength{\tabcolsep}{5pt}
\begin{tabularx}{\textwidth}{lXXc}
\toprule
\textbf{Game}
&
\textbf{Exploration}
&
\textbf{Evaluation}
&
\textbf{Max steps}
\tabularnewline
\midrule

Chess
&
$15\times15$ board, 12 pieces, and a 7-step history.
&
Same rules as exploration but 22 pieces.
&
5
\tabularnewline

Minesweeper
&
$9\times9$ board with 10--20 mines and two extra lives.
&
The first mine hit terminates the episode.
&
64
\tabularnewline

Nullify
&
Expressions generated with 3--7 construction steps.
&
Expressions generated with 5--10 construction steps.
&
50
\tabularnewline

Plants-vs-Zombies
&
$3\times7$ battlefield.
&
$5\times6$ battlefield.
&
64
\tabularnewline

Snake
&
Collisions and invalid actions are treated as no-ops.
&
Collisions and invalid actions terminate the episode.
&
64
\tabularnewline

Tetris
&
$10\times10$ board.
&
$8\times8$ board.
&
64
\tabularnewline

Trust Evolution
&
Opponent sampling is biased toward easier strategies.
&
Opponent strategies are sampled uniformly.
&
10
\tabularnewline

\bottomrule
\end{tabularx}
\caption{
Exploration and evaluation configurations for the seven S$^3$Gym games.
Exploration generally uses more permissive rules, while evaluation
uses stricter or more difficult configurations.
}
\label{tab:env-setting}
\end{table}
\FloatBarrier

\subsection{Experimental Setup}
\label{sec:setting}

\paragraph{Models.}
We evaluate seven representative proprietary LLMs for which all seven game results are complete: GPT-4o, GPT-4.1, o3-mini, Gemini-2.5-Flash, Gemini-2.5-Pro, GPT-5.5, and Gemini-3.5-Flash. Models with unfinished checkpoints are excluded from the main table to avoid comparisons based on different game coverage.

\paragraph{Context-level pathways.}
We compare two non-parametric self-improvement pathways. \textbf{History ICL} directly appends the interaction history to the context, whereas \textbf{Summary Memory} compresses previous trajectories into model-generated rules, mistakes, and future directions. The former preserves detailed evidence but requires a larger context budget; the latter imposes an information bottleneck and therefore depends more strongly on the quality of Self-Judging and summarization.

\begin{table}[H]
\centering
\small
\begin{tabularx}{0.6\columnwidth}{lX}
\toprule
\textbf{Setting} & \textbf{Value} \tabularnewline
\midrule
History-ICL input limit & 100,000 tokens \tabularnewline
Output generation cap & 14,000 new tokens per response \tabularnewline
Controller interaction cap & 64 steps per episode \tabularnewline
Exploration budget & 30 exploration episodes \tabularnewline
Evaluation interval & Every 3 exploration episodes \tabularnewline
Evaluation set size & 3 strict-mode episodes per checkpoint \tabularnewline
Evaluation checkpoints & $x\in\{0,3,6,\ldots,30\}$ \tabularnewline
\bottomrule
\end{tabularx}
\caption{Runtime configuration for the main context-level experiments. Game-specific horizons smaller than 64 override the controller-level step cap.}
\label{tab:runtime-setting}
\end{table}

\paragraph{Metrics.}
Let $y_{m,p,g,k}$ denote the evaluation score of model $m$, pathway $p$, and game $g$ at checkpoint $x_k$. We report three complementary metrics for every model--game pair:

\begin{align}
\operatorname{Avg}_{m,p,g}
&= \frac{1}{K+1}\sum_{k=0}^{K} y_{m,p,g,k}, \\
\operatorname{Max}_{m,p,g}
&= \max_{0\leq k\leq K} y_{m,p,g,k}, \\
\operatorname{AUC}^{+}_{m,p,g}
&= \int_{0}^{30}\max\!\left(0,y_{m,p,g}(x)-y_{m,p,g,0}\right)\,dx.
\label{eq:experiment_metrics}
\end{align}

The continuous curve $y(x)$ in $\operatorname{AUC}^{+}$ is obtained through piecewise-linear interpolation between adjacent checkpoints. Avg. measures overall performance throughout exploration; Max. captures the best observed capability, and $\operatorname{AUC}^{+}$ measures sustained improvement above the initial evaluation score. Because the games use different score scales, raw $\operatorname{AUC}^{+}$ values should be compared \emph{within the same game}, rather than directly across games.

\FloatBarrier
\subsection{Main Results}
\label{sec:main_results}

Table~\ref{tab:main_results} reports Avg., Max., and $\operatorname{AUC}^{+}$ for all complete models under History ICL and Summary Memory. Bold values indicate the best result for each metric within the corresponding pathway and game.

\paragraph{Overall context-level performance.}
The results show that self-improvement potential is jointly determined by the base model, the experience pathway, and the environment. Under History ICL, Gemini-3.5-Flash obtains the strongest Chess results and the largest AUC on Snake and Trust Evolution, whereas GPT-5.5 leads the more brittle Minesweeper, Nullify, and Tetris tasks and achieves the largest PvZ AUC. Under Summary Memory, GPT-5.5 becomes strongest on Chess, Nullify, and Tetris and reaches the highest average and maximum scores on Snake. Gemini-3.5-Flash remains strongest in average performance on PvZ and Trust Evolution. These results indicate that high base capability alone does not determine self-improvement; models also differ in how effectively they retrieve, compress, and apply their own experience.

\paragraph{History ICL versus Summary Memory.}
Summary Memory is not uniformly superior to raw history. It can substantially improve models for which long trajectories are difficult to use directly. For example, Gemini-2.5-Flash increases its Minesweeper $\operatorname{AUC}^{+}$ from $0.000$ under ICL to $7.794$ with Summary Memory, and its PvZ $\operatorname{AUC}^{+}$ from $24.402$ to $238.501$. GPT-5.5 similarly improves its Chess $\operatorname{AUC}^{+}$ from $0.474$ to $16.840$. However, summarization may discard state-specific evidence or preserve incorrect judgments. GPT-5.5's PvZ $\operatorname{AUC}^{+}$ decreases from $548.499$ to $33.219$, while Gemini-3.5-Flash decreases from $12.280$ to $0.000$ on Chess and

\begin{table*}[!p]
\centering
\caption{
Main results on seven S$^3$Gym games. For each model--game pair,
Avg. is the mean score over all available evaluation checkpoints,
Max. is the maximum score, and $\mathrm{AUC}^{+}$ is the positive
area above the initial-score baseline. Higher is better for all
metrics. Explicitly marked concurrency failures are excluded.
}
\label{tab:main_results}

\begingroup
\fontsize{8}{8.8}\selectfont
\renewcommand{\arraystretch}{0.93}
\setlength{\tabcolsep}{2.5pt}
\setlength{\aboverulesep}{1pt}
\setlength{\belowrulesep}{1pt}
\begin{tabular*}{\textwidth}{@{\extracolsep{\fill}}l*{7}{r}@{}}
\toprule
\multicolumn{8}{l}{\textbf{(a) Average score (Avg.)}} \\
\midrule
\textbf{Model} & \textbf{Chess} & \textbf{Minesweeper} & \textbf{Nullify}
& \textbf{Tetris} & \textbf{Snake} & \textbf{PvZ}
& \shortstack{\textbf{Trust}\textbf{Evo}} \\
\midrule
\multicolumn{8}{c}{\textbf{ICL (History Context)}} \\
\midrule
GPT-4o           & 0.007 & 0.044 & 0.030 & 0.030 & 0.212 & 11.636 & 8.242 \\
GPT-4.1          & 0.010 & 0.042 & 0.030 & 0.152 & 0.515 & 16.909 & 7.371 \\
o3-mini          & 0.012 & 0.222 & 0.030 & 1.273 & 4.000 & 26.333 & 9.242 \\
Gemini-2.5-Flash & 0.009 & 0.063 & 0.030 & 0.121 & 1.424 & 20.697 & 9.008 \\
Gemini-2.5-Pro   & 0.025 & 0.421 & 0.212 & 0.273 & 1.576 & 28.242 & 14.500 \\
GPT-5.5          & 0.017 & \textbf{0.604} & \textbf{0.545} & \textbf{4.242}
                 & \textbf{9.061} & 44.033 & 7.848 \\
Gemini-3.5-Flash & \textbf{0.547} & 0.482 & 0.394 & 1.970 & 8.455
                 & \textbf{44.697} & \textbf{17.242} \\
\addlinespace[1pt]
\midrule
\multicolumn{8}{c}{\textbf{Summary Memory}} \\
\midrule
GPT-4o           & 0.005 & 0.013 & 0.000 & 0.030 & 0.303 & 11.818 & 13.947 \\
GPT-4.1          & 0.010 & 0.012 & 0.030 & 0.182 & 0.545 & 14.727 & 13.371 \\
o3-mini          & 0.006 & 0.425 & 0.061 & 1.788 & 6.727 & 26.424 & 7.174 \\
Gemini-2.5-Flash & 0.074 & 0.438 & 0.091 & 0.152 & 1.364 & 21.939 & 8.720 \\
Gemini-2.5-Pro   & 0.110 & 0.466 & 0.182 & 0.545 & 1.273 & 19.667 & 9.568 \\
GPT-5.5          & \textbf{0.533} & \textbf{0.742} & \textbf{0.576}
                 & \textbf{3.515} & \textbf{10.697} & 33.500 & 8.697 \\
Gemini-3.5-Flash & 0.208 & 0.603 & 0.455 & 1.091 & 10.242
                 & \textbf{43.879} & \textbf{14.394} \\

\midrule
\multicolumn{8}{l}{\textbf{(b) Maximum score (Max.)}} \\
\midrule
\textbf{Model} & \textbf{Chess} & \textbf{Minesweeper} & \textbf{Nullify}
& \textbf{Tetris} & \textbf{Snake} & \textbf{PvZ}
& \shortstack{\textbf{Trust}\textbf{Evo}} \\
\midrule
\multicolumn{8}{c}{\textbf{ICL (History Context)}} \\
\midrule
GPT-4o           & 0.018 & 0.131 & 0.333 & 0.333 & 0.667 & 17.667 & 11.500 \\
GPT-4.1          & 0.018 & 0.189 & 0.333 & 0.667 & 1.667 & 21.667 & 10.250 \\
o3-mini          & 0.022 & 0.500 & 0.333 & 2.667 & 9.667 & 29.333 & 13.250 \\
Gemini-2.5-Flash & 0.022 & 0.670 & 0.333 & 0.333 & 2.333 & 29.000 & 13.583 \\
Gemini-2.5-Pro   & 0.089 & 0.659 & 0.667 & 1.000 & 4.333 & 33.000 & 19.000 \\
GPT-5.5          & 0.067 & \textbf{0.974} & \textbf{1.000} & \textbf{11.333}
                 & 11.333 & 48.000 & \textbf{23.000} \\
Gemini-3.5-Flash & \textbf{0.742} & 0.705 & \textbf{1.000} & 3.000
                 & \textbf{13.000} & \textbf{62.667} & 21.000 \\
\addlinespace[1pt]
\midrule
\multicolumn{8}{c}{\textbf{Summary Memory}} \\
\midrule
GPT-4o           & 0.018 & 0.033 & 0.000 & 0.333 & 1.000 & 20.333 & 19.333 \\
GPT-4.1          & 0.031 & 0.083 & 0.333 & 0.667 & 1.000 & 19.333 & 15.917 \\
o3-mini          & 0.018 & 0.608 & 0.333 & 4.000 & 10.333 & 29.000 & 10.083 \\
Gemini-2.5-Flash & 0.147 & 0.665 & 0.333 & 0.667 & 2.333 & 26.333 & 13.750 \\
Gemini-2.5-Pro   & 0.258 & 0.974 & 0.333 & 2.000 & 3.000 & 26.667 & 14.333 \\
GPT-5.5          & \textbf{0.773} & \textbf{1.000} & \textbf{1.000}
                 & \textbf{6.667} & \textbf{14.000} & 41.333 & 17.000 \\
Gemini-3.5-Flash & 0.524 & 0.961 & \textbf{1.000} & 2.333 & 12.000
                 & \textbf{51.667} & \textbf{20.333} \\

\midrule
\multicolumn{8}{l}{\textbf{(c) Positive area above baseline ($\mathrm{AUC}^{+}$)}} \\
\midrule
\textbf{Model} & \textbf{Chess} & \textbf{Minesweeper} & \textbf{Nullify}
& \textbf{Tetris} & \textbf{Snake} & \textbf{PvZ}
& \shortstack{\textbf{Trust}\textbf{Evo}} \\
\midrule
\multicolumn{8}{c}{\textbf{ICL (History Context)}} \\
\midrule
GPT-4o           & 0.233 & 1.437 & 0.500 & 1.000 & 0.000 & 0.052 & 3.443 \\
GPT-4.1          & 0.186 & 1.329 & 1.000 & 0.000 & 8.501 & 129.000 & 5.966 \\
o3-mini          & 0.004 & 0.000 & 1.000 & 0.000 & 0.000 & 5.383 & 62.638 \\
Gemini-2.5-Flash & 0.017 & 0.000 & 1.000 & 0.000 & 7.317 & 24.402 & 6.251 \\
Gemini-2.5-Pro   & 0.000 & \textbf{3.539} & \textbf{1.500} & 0.000
                 & 0.000 & 60.416 & 161.876 \\
GPT-5.5          & 0.474 & 0.959 & 1.000 & \textbf{96.251} & 0.163
                 & \textbf{548.499} & 164.411 \\
Gemini-3.5-Flash & \textbf{12.280} & 1.217 & 0.417 & 34.000
                 & \textbf{37.387} & 161.945 & \textbf{200.000} \\
\addlinespace[1pt]
\midrule
\multicolumn{8}{c}{\textbf{Summary Memory}} \\
\midrule
GPT-4o           & 0.007 & 0.046 & 0.000 & 0.500 & 1.917 & 12.778 & 10.350 \\
GPT-4.1          & 0.177 & 0.046 & 1.000 & 0.000 & 0.000 & 75.462 & 154.876 \\
o3-mini          & 0.079 & 0.001 & 2.000 & 28.598 & 0.300 & 40.425 & 30.521 \\
Gemini-2.5-Flash & 2.147 & \textbf{7.794} & 3.000 & 0.500 & 0.000
                 & \textbf{238.501} & 18.443 \\
Gemini-2.5-Pro   & 0.573 & 0.721 & 0.000 & 10.084 & \textbf{11.321}
                 & 11.872 & 10.624 \\
GPT-5.5          & \textbf{16.840} & 2.951 & \textbf{7.418}
                 & \textbf{50.801} & 0.000 & 33.219 & \textbf{240.500} \\
Gemini-3.5-Flash & 0.000 & 0.793 & 5.584 & 15.808 & 0.750 & 123.535 & 55.095 \\
\bottomrule
\end{tabular*}
\endgroup

\vspace{2pt}
\begin{minipage}{0.98\textwidth}
\fontsize{7.4}{8.5}\selectfont
\emph{Note.}
$\mathrm{AUC}^{+}$ is computed over checkpoints
$x\in\{0,3,\ldots,30\}$ using piecewise-linear trapezoidal
integration of
$\int_{0}^{30}\max\!\left(0,y(x)-y_0\right)\,dx$.
Means and maxima exclude explicitly marked invalid concurrency points;
for $\mathrm{AUC}^{+}$, adjacent valid checkpoints are connected.
Models with unfinished results in any game are omitted.
\end{minipage}
\end{table*}
\FloatBarrier

% Paired single-column trend figures.
% Required package in the preamble:
% \usepackage{caption}
%
% The paths below are resolved relative to the root TeX file. Direct
% \includegraphics calls are intentional: a missing file now raises a clear
% compilation error instead of silently producing an empty placeholder.

\newcommand{\sThreeSingleTrendPair}[1]{%
  \noindent
  \begin{minipage}[t]{0.44\linewidth}
    \centering
    \includegraphics[width=\linewidth]{pics/trends/icl_#1.pdf}
  \end{minipage}%
  \hfill
  \begin{minipage}[t]{0.44\linewidth}
    \centering
    \includegraphics[width=\linewidth]{pics/trends/summary_#1.pdf}
  \end{minipage}%
}

\begin{figure}[!htbp]
\centering
\sThreeSingleTrendPair{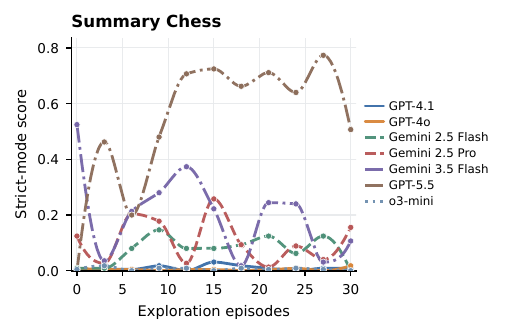}
\vspace{2pt}

\sThreeSingleTrendPair{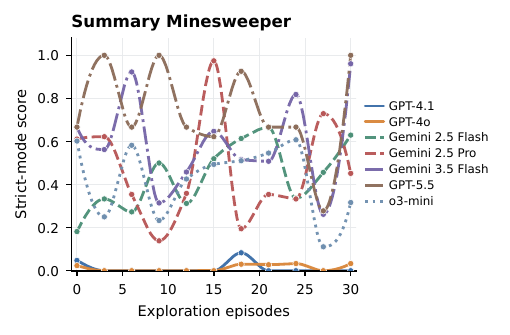}

\vspace{-3pt}
\caption{
Paired History-ICL and Summary-Memory trajectories for Chess and
Minesweeper. The horizontal axis denotes exploration episodes, and
the vertical axis denotes strict-mode evaluation performance. Within
each game, the two panels use the same vertical scale.
}
\label{fig:context-trends-a}
\end{figure}

\begin{figure}[!htbp]
\centering
\sThreeSingleTrendPair{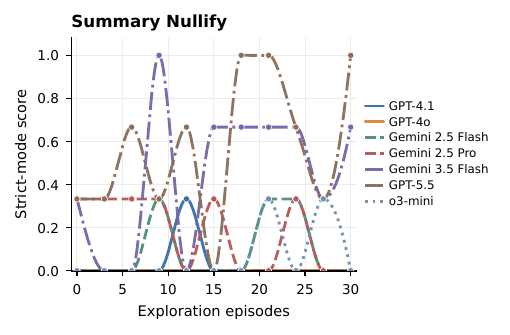}

\vspace{2pt}

\sThreeSingleTrendPair{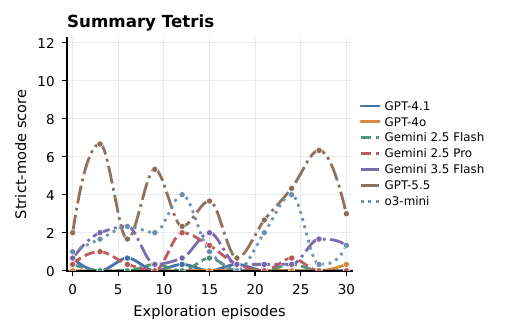}

\vspace{-3pt}
\caption{
Paired History-ICL and Summary-Memory trajectories for Nullify and
Tetris. Model colors and line styles are shared across all panels.
}
\label{fig:context-trends-b}
\end{figure}

\begin{figure}[!htbp]
\centering
\sThreeSingleTrendPair{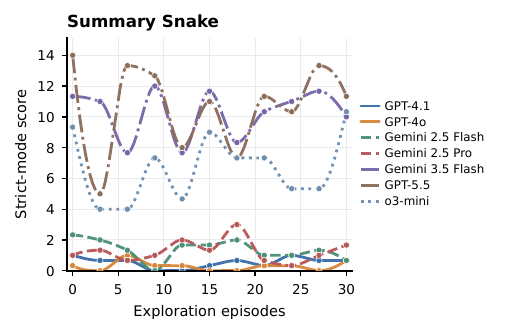}

\vspace{2pt}

\sThreeSingleTrendPair{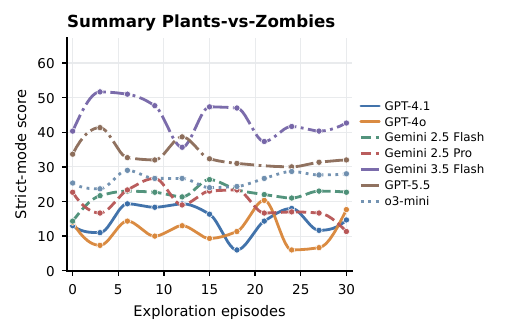}

\vspace{-3pt}
\caption{
Paired History-ICL and Summary-Memory trajectories for Snake and
Plants-vs-Zombies. Within each game, both pathways use the same
vertical scale.
}
\label{fig:context-trends-c}
\end{figure}

\begin{figure}[!htbp]
\centering
\sThreeSingleTrendPair{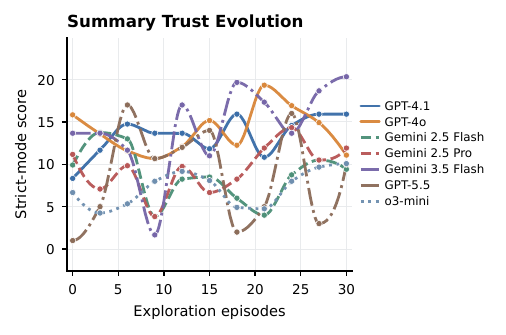}

\vspace{-3pt}
\caption{
Paired History-ICL and Summary-Memory trajectories for Trust Evolution.
}
\label{fig:context-trends-d}
\end{figure}

 from $200.000$ to $55.095$ on Trust Evolution. Summary Memory should therefore be viewed as a selective compression mechanism rather than an unconditional improvement over History ICL.
 
\paragraph{Cross-game consistency.}
The seven games reveal different self-improvement regimes. Chess, Minesweeper, and Nullify have sparse or highly discrete scores, so a single successful episode can strongly affect Max. while producing only a small Avg. or AUC. Tetris, Snake, PvZ, and Trust Evolution provide denser sequential feedback and expose clearer differences in sustained adaptation. The three metrics are consequently complementary: Max. identifies occasional capability breakthroughs, Avg. reflects overall performance, and $\operatorname{AUC}^{+}$ distinguishes persistent improvement from isolated peaks. Since score scales differ substantially across games, cross-game consistency should be assessed through within-game ranks or the number of games improved, rather than by averaging raw AUC magnitudes.
\section{Analysis}
\label{sec:analysis}

\subsection{RQ1: Can Parameter Training Enable Self-Improvement?}
\label{sec:rq1_training}

RQ1 examines whether an agent can internalize its interaction experience through parameter updates and subsequently improve without access to historical trajectories or external memory at inference time. Compared with context-level adaptation, parameter training offers a more persistent improvement pathway, but it also carries greater risk: noisy or incorrectly judged trajectories may be consolidated into the model and overwrite previously effective behavior.

We evaluate Qwen3-8B at 20 consecutive checkpoints, with epoch 0 representing the original model and epochs 1--19 representing models updated using exploration trajectories. At each checkpoint, the model is evaluated on the strict configurations of all seven S$^3$Gym games without History ICL or Summary Memory. We report the post-training average and maximum scores, the final change
$\Delta_{\mathrm{Final}}=y_{19}-y_0$, and the positive area above the initial baseline,
\[
\operatorname{AUC}^{+}
=
\int_{0}^{19}
\max\!\left(0,y(e)-y_0\right)\,de,
\]
where $y(e)$ is obtained by linearly interpolating adjacent checkpoints.

\begin{figure}[h]
    \centering
    \includegraphics[
        width=0.88\textwidth
    ]{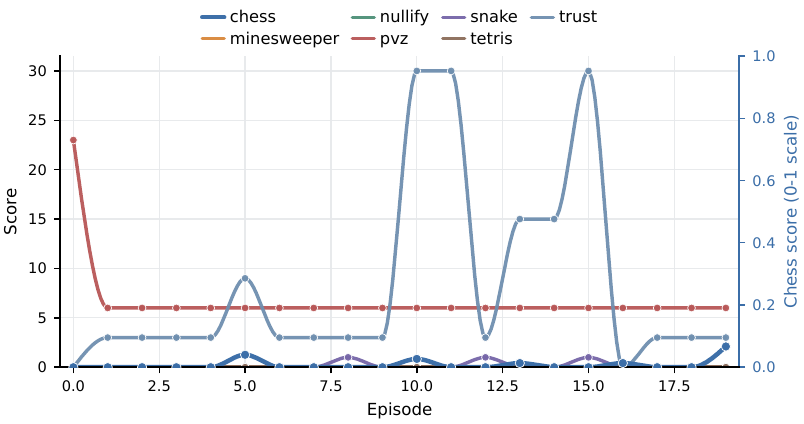}
    \caption{
    Strict-mode evaluation performance of Qwen3-8B over 20
    training checkpoints. Epoch 0 denotes the original model.
    Because the games use different score scales, each trajectory
    should be interpreted relative to its own initial performance.
    }
    \label{fig:training_qwen3}
\end{figure}

\begin{itemize}
    \item \textbf{Training yields substantial self-improvement on Trust Evolution.}
    The score rises from 0 to a maximum of 30 and remains above the initial baseline at 18 of the 19 updated checkpoints. The post-training average reaches 8.684 and $\operatorname{AUC}^{+}$ reaches 163.5, providing the clearest evidence that interaction-derived strategies can be internalized into model parameters.

    \item \textbf{The gains on Chess and Snake are limited or transient.}
    Chess remains at zero for most checkpoints but reaches 0.0667 at the final checkpoint, indicating a sparse yet positive gain. Snake reaches a score of 1 at epochs 8, 12, and 15 but subsequently returns to zero. Training therefore discovers occasional successful behavior without retaining it consistently.

    \item \textbf{Minesweeper, Nullify, and Tetris show no measurable improvement.}
    Their evaluation scores remain zero throughout training. One possible explanation is that exploration produces too few successful trajectories to bootstrap an effective supervised signal, although verifying this hypothesis requires trajectory-level analysis.

    \item \textbf{Training produces clear negative transfer on Plants-vs-Zombies.}
    The initial score is 23, whereas every updated checkpoint obtains a score of 6. This persistent degradation indicates that parameter updates can overwrite previously effective behavior. Potential causes include overfitting to exploration trajectories, mismatch between relaxed exploration and strict evaluation, or unreliable Self-Judging signals.

    \item \textbf{Overall, parameter training enables self-improvement, but not consistently across tasks.}
    Qwen3-8B improves substantially on Trust Evolution, weakly on Chess, and intermittently on Snake, while showing no improvement on three games and severe degradation on Plants-vs-Zombies. Effective training-level self-improvement therefore depends not only on the ability to update parameters, but also on the quality, diversity, and correctness of the experience being internalized.
\end{itemize}

\subsection{RQ2: Is Agent Self-Judging Reliable?}
\label{sec:rq2-self-judging}

Self-Judging is a central component of S$^3$Gym because the model-generated score $s_i$ is stored with each transition and subsequently influences trajectory selection, summary construction, and training-data generation. If $s_i$ is misaligned with the environment signal, the agent may preserve harmful actions as useful experience or discard transitions that should have been retained. We therefore evaluate Self-Judging as an independent capability and examine whether judgment quality predicts subsequent improvement.

\paragraph{Step-level reliability.}

For each transition, we compare the model-generated score $s_i$ with the environment reward $r_i$. We first measure event agreement, which indicates whether the model and environment agree on the presence of a positive outcome:

\begin{equation}
\mathrm{Agree}
=
\frac{1}{N}
\sum_{i=1}^{N}
\mathbf{1}
\left[
\mathbf{1}(s_i>0)
=
\mathbf{1}(r_i>0)
\right].
\end{equation}

Because reward scales differ across games, we additionally report normalized mean absolute error:

\begin{equation}
\mathrm{NMAE}
=
\frac{1}{N}
\sum_{i=1}^{N}
\left|
\frac{s_i-s_{\min}}{s_{\max}-s_{\min}}
-
\frac{r_i-r_{\min}}{r_{\max}-r_{\min}}
\right|.
\end{equation}

We also measure over-confidence,
$\Pr(s_i>0,r_i\leq0)$, and under-confidence,
$\Pr(s_i\leq0,r_i>0)$. The analysis covers the seven target models and games under both direct-history and summary-memory conditions, comprising 98 runs and 116,117 transitions.

\begin{table}[H]
\centering
\small
\setlength{\tabcolsep}{4pt}
\begin{tabular}{lrrrrr}
\toprule
\textbf{Game}

&
\textbf{Agreement}
&
\textbf{NMAE}
&
\textbf{Over-conf.}
&
\textbf{Under-conf.}
\tabularnewline
\midrule
Chess       &  0.496 & 0.141 & 0.365 & 0.139 \tabularnewline
Minesweeper &  0.820 & 0.524 & 0.062 & 0.118 \tabularnewline
Nullify     &  0.827 & 0.056 & 0.170 & 0.004 \tabularnewline
PvZ         &  0.881 & 0.882 & 0.032 & 0.087 \tabularnewline
Snake       &  0.879 & 0.121 & 0.110 & 0.011 \tabularnewline
Tetris      &  0.846 & 0.041 & 0.151 & 0.002 \tabularnewline
Trust       &  0.665 & 0.392 & 0.291 & 0.044 \tabularnewline
\bottomrule
\end{tabular}
\caption{
Step-level reliability of model Self-Judging.
Agreement measures binary agreement on whether a transition receives
a positive score. NMAE measures score calibration after game-level
min--max normalization.
}
\label{tab:self-judging-alignment}
\end{table}

Table~\ref{tab:self-judging-alignment} shows that Self-Judging is only partially reliable. PvZ, Snake, Tetris, Minesweeper, and Nullify exhibit relatively high event agreement, but this result is partly explained by the large proportion of zero-reward transitions. NMAE reveals whether the magnitude of the model's judgment is also well calibrated. Tetris and Nullify achieve low NMAE, whereas PvZ exhibits the largest calibration error despite its high binary agreement. Models can therefore recognize whether an action appears locally useful while substantially misestimating its value.

Chess and Trust are the clearest failure cases. Chess has near-random event agreement and the highest over-confidence rate. A prediction may be syntactically complete while still misplacing several pieces, causing the model's confidence in its response to diverge from the environment score. Trust similarly exhibits weak agreement, high NMAE, and substantial over-confidence. These results suggest that Self-Judging becomes less reliable when reward depends on hidden dynamics, opponent strategy, or exact multi-object state prediction rather than an immediately observable local consequence.

\paragraph{Judgment--improvement coupling.}

Reliable local judgments are useful only if the agent can convert them into better subsequent behavior. To test this relationship, we divide exploration trajectories into blocks $B_t$ between adjacent evaluation checkpoints. For each block, we calculate event agreement $A_t$, normalized calibration error $E_t$, and the normalized score change at the following evaluation:

\begin{equation}
g_t
=
\frac{
y_t-y_{t-3}
}{
\max_{g,\tau}y_{g,\tau}
-
\min_{g,\tau}y_{g,\tau}
}.
\end{equation}

We then measure $\rho(A_t,g_t)$ and $\rho(-E_t,g_t)$, together with the difference in average score gain between the highest- and lowest-quality judgment tertiles.

\begin{table}[H]
\centering
\small
\setlength{\tabcolsep}{4pt}
\begin{tabular}{lrrrr}
\toprule
\textbf{Game}

&
$\boldsymbol{\rho(A,g)}$
&
$\boldsymbol{\rho(-E,g)}$
&
$\boldsymbol{\mathrm{Gap}_A}$
&
$\boldsymbol{\mathrm{Gap}_E}$
\tabularnewline
\midrule
All         &  -0.010 & -0.018 & -0.004 & -0.016 \tabularnewline
Chess       &  -0.014 & -0.058 &  0.028 & -0.039 \tabularnewline
Minesweeper &   0.071 &  0.006 &  0.069 &  0.005 \tabularnewline
Nullify     &  -0.032 & -0.082 & -0.065 & -0.058 \tabularnewline
PvZ         &  -0.038 & -0.023 &  0.014 &  0.010 \tabularnewline
Snake       &   0.022 &  0.022 &  0.009 &  0.009 \tabularnewline
Tetris      &  -0.048 & -0.021 &  0.010 &  0.004 \tabularnewline
Trust       &  -0.014 & -0.003 & -0.001 &  0.006 \tabularnewline
\bottomrule
\end{tabular}
\caption{
Blockwise judgment--improvement coupling.
$A$ is event agreement, $E$ is normalized calibration error, and
$g$ is the normalized strict-mode score change at the subsequent
evaluation. The gap metrics compare the highest and lowest judgment-
quality tertiles.
}
\label{tab:self-judging-coupling}
\end{table}

The coupling results are close to zero across the full benchmark:
$\rho(A,g)=-0.010$ and $\rho(-E,g)=-0.018$. The corresponding tertile gaps are also small and slightly negative. Minesweeper exhibits a weak positive relationship between event agreement and subsequent score gain, but its calibration coupling remains negligible. Overall, accurate local scoring does not reliably produce improvement at the next evaluation checkpoint.

The run-level analysis leads to the same conclusion. To quantify sustained improvement while accounting for differences in initial performance, we use the \textbf{normalized above-baseline area (NABA)}, defined as

$$
\mathrm{NABA}
=
\frac{\mathrm{AUC}^{+}}{\max(y_0,\epsilon)},
$$

where $y_0$ is the initial evaluation score and $\epsilon$ denotes the first non-zero score scale when $y_0=0$. Event agreement exhibits only weak negative correlations with NABA under both Pearson and Spearman measures ($r=-0.23$ and $\rho=-0.11$), while the positive-reward and positive-self-score rates show only weak positive correlations. These results indicate that accurate local \textbf{Self-Judging} alone is insufficient for \textbf{Self-Improvement}: agents must also transform feedback into reusable state abstractions, decision rules, and exploration strategies.

\subsection{RQ3: Do Score-Conditioned Summaries Produce Effective Iteration Directions?}
\label{sec:rq3-summary}

Summary Memory imposes a stronger requirement than direct history reuse. Rather than merely conditioning on previous trajectories, the model must identify which low-scoring behaviors should be suppressed, which high-scoring behaviors should be retained, and how these observations should be converted into an executable strategy for future episodes. We evaluate this capability by comparing score-conditioned summaries with direct-history ICL and by inspecting representative successful and unsuccessful summaries.

\paragraph{Summary Memory versus direct history.}

For every model--game pair, we calculate
$\Delta\mathrm{NABA}$ as Summary-Memory NABA minus Direct-History NABA. Positive values indicate that summarization produces a more effective improvement trajectory than retaining the raw interaction history.

\begin{table}[H]
\centering
\small
\setlength{\tabcolsep}{4pt}
\begin{tabular}{lrrrrr}
\toprule
\textbf{Game}
&
\textbf{Pairs}
&
\textbf{Summary wins}
&
\textbf{History wins}
&
\textbf{Ties}
&
\textbf{Mean $\boldsymbol{\Delta}$NABA}
\tabularnewline
\midrule
Chess       & 7 & 3 & 3 & 1 &  3.37 \tabularnewline
Minesweeper & 7 & 3 & 4 & 0 & -1.12 \tabularnewline
Nullify     & 7 & 4 & 2 & 1 &  2.79 \tabularnewline
PvZ         & 7 & 3 & 4 & 0 & -2.11 \tabularnewline
Snake & 7 & 3 & 4 & 0 & -0.81  \tabularnewline
Tetris      & 7 & 5 & 1 & 1 &  6.20 \tabularnewline
Trust       & 7 & 4 & 3 & 0 &  8.43 \tabularnewline
\bottomrule
\end{tabular}
\caption{
Summary Memory compared with direct-history ICL.
Each pair evaluates the same model and game under the two
history-organization strategies.
}
\label{tab:summary-vs-history}
\end{table}

Summary Memory improves average NABA on Nullify, Tetris, and Trust. These environments admit compact and reusable abstractions, such as planning backward toward exact cancellation, preserving a stable board surface, or adapting behavior to an inferred opponent strategy. In such cases, summaries remove redundant trajectory details while retaining the information most relevant to future decisions.

In contrast, direct history performs better on average in Minesweeper, PvZ, and Snake. These games depend heavily on local board configurations, lane timing, hazard positions, and partially observed risk. A summary may preserve a sensible high-level objective while discarding the state-contingent details required to execute it correctly. Summary Memory is therefore most effective when experience can be compressed into a stable policy rule, rather than when success depends on reconstructing the exact current state.

\paragraph{Trajectory-level evidence.}

The following cases illustrate when score-conditioned summaries become executable iteration directions and when they remain too generic to improve behavior.

\definecolor{rqpos}{RGB}{34,121,104}
\definecolor{rqneg}{RGB}{174,74,74}

\newcommand{\rqcasecard}[6]{%
\begin{tcolorbox}[
    enhanced,
    colback=#1!4!white,
    colframe=#1!70!black,
    boxrule=0.6pt,
    arc=1.2mm,
    left=1.7mm,
    right=1.7mm,
    top=1.2mm,
    bottom=1.2mm,
    title={#2},
    coltitle=white,
    colbacktitle=#1!75!black,
    fonttitle=\bfseries\footnotesize,
    before skip=4pt,
    after skip=6pt
]
\footnotesize
\textbf{Summary direction.} #3\par\vspace{2pt}
\textbf{Trajectory evidence.} #4\par\vspace{2pt}
\textbf{Outcome.} #5\par\vspace{2pt}
\textbf{Takeaway.} #6
\end{tcolorbox}
}

\subsubsection*{Successful summaries}

\noindent
\begin{minipage}[t]{0.49\textwidth}
\rqcasecard
{rqpos}
{GPT-5.5 / Trust}
{Always defect when the observed payoff structure consistently favors defection; avoid unnecessary reciprocal exploration.}
{At the first decision, the agent selected \texttt{cheat} and received an environment reward of 3.0.}
{$\Delta\mathrm{NABA}=+66.89$.}
{The summary translates payoff feedback into an explicit opponent-conditioned policy.}
\end{minipage}
\hfill
  \begin{minipage}[t]{0.49\textwidth}
  \rqcasecard
  {rqpos}
  {o3-mini / Tetris}
  {Jointly inspect the current and next blocks, test alternative rotations, prioritize line-clearing placements, and maintain a low surface without isolated holes.}
  {In episode 10, the agent cleared four lines. It selected \texttt{8 0} twice, with both its self-score and the environment reward equal to 1.0 on each move.}
  {$\Delta\mathrm{NABA}=+14.42$.}
  {The summary converts successful line clears into an executable look-ahead rule.}
  \end{minipage}

\noindent
\begin{minipage}[t]{0.49\textwidth}
\rqcasecard
{rqpos}
{Gemini-2.5-Flash / Minesweeper}
{Apply local number constraints: expand zero regions, flag forced mines, and uncover cells that are provably safe.}
{The agent selected \texttt{flag (2,2)} on a partially revealed board and received a reward of 0.0625.}
{$\Delta\mathrm{NABA}=+25.49$.}
{Summaries help when they express local reasoning as executable constraint rules.}
\end{minipage}
\hfill
\begin{minipage}[t]{0.49\textwidth}
\rqcasecard
{rqpos}
{Gemini-3.5-Flash / Nullify}
{Plan backward to create exact opposites, use floor or ceiling operations to remove decimals, and track index changes after every merge.}
{With terminal units \texttt{-19} and \texttt{19}, the agent selected \texttt{0 1} and received a reward of 1.0.}
{$\Delta\mathrm{NABA}=+8.50$.}
{The summary succeeds because it specifies a concrete arithmetic planning procedure.}
\end{minipage}

\subsubsection*{Failure summaries}

\noindent
\begin{minipage}[t]{0.49\textwidth}
\rqcasecard
{rqneg}
{GPT-5.5 / PvZ}
{Prioritize sun production, early defense, Wall-nut stalling, and coverage of threatened lanes.}
{The agent opened with \texttt{X 1 0} and received a reward of 1.0, but its later strict score fell from 38.67 to 32.33.}
{$\Delta\mathrm{NABA}=-2.56$.}
{The advice is reasonable but too coarse to capture lane timing, current zombie positions, and the available sun budget.}
\end{minipage}
\hfill
\begin{minipage}[t]{0.49\textwidth}
\rqcasecard
{rqneg}
{Gemini-3.5-Flash / Chess}
{Track the movement rule of every piece and verify all coordinates before submitting the answer.}
{In a low-scoring episode, the model produced \texttt{END} and received zero total reward despite having generated a detailed rule summary.}
{$\Delta\mathrm{NABA}=-33.27$.}
{A high-level reminder does not resolve the exact multi-piece localization problem.}
\end{minipage}

\noindent
\begin{minipage}[t]{0.49\textwidth}
\rqcasecard
{rqneg}
{GPT-4.1 / Minesweeper}
{Begin from corners, expand zero regions, flag only certain mines, and stop when no safe progress is available.}
{The agent selected \texttt{uncover (8,0)} after exhausting its available lives, and the subsequent checkpoint score fell from 0.0606 to 0.}
{$\Delta\mathrm{NABA}=-22.09$.}
{Generic heuristics do not substitute for exact constraint propagation.}
\end{minipage}
\hfill
\begin{minipage}[t]{0.49\textwidth}
\rqcasecard
{rqneg}
{GPT-4o / Snake}
{Move toward food while avoiding walls, obstacles, the snake body, and direct reversals.}
{Given a board containing food, obstacles, and its own body, the agent selected \texttt{DOWN}; it received no reward, and the strict score fell from 1.0 to 0.}
{$\Delta\mathrm{NABA}=-22.00$.}
{The summary omits the precise geometry required for safe navigation.}
\end{minipage}

Taken together, these cases reveal two requirements for effective summary-based improvement. First, the score must identify a meaningful and reusable failure mode rather than merely indicate that the episode ended poorly. Second, the summary must convert that signal into an action rule precise enough to execute in a new state. Summary Memory is therefore a selective improvement mechanism: it is effective when experience admits compact causal abstractions, but it cannot universally replace direct access to state-rich interaction histories.

\section{Conclusion}

S3GYM turns self-improvement from a broad claim into a measurable agent capability. By separating relaxed exploration from stricter evaluation, and by supporting both context-level and training-level reuse of trajectories, the benchmark tests whether LLMs can transform interaction history into future performance. The evidence so far is mixed in a useful way. Summary memory helps in games where compact rules transfer across episodes, while direct history is often better in reactive control games where details matter. Training on self-generated trajectories remains unstable in the current implementation. These findings suggest that future self-improving agents need better judgment calibration, memory selection, and trajectory filtering, not just more interaction data.
\section{Contributions}

\textbf{Core Contributors} \\
Jiajun Shi, Siyuan Tao, Yuhao Wu, Zexuan Wang, Jingyuan Zhang

\vspace{0.25em}

\textbf{Contributors} \\
Jiaheng Liu, Xinping Lei, Xinrong Zhang, Siyuan Fang, Zhewen Tan,
Tianle Cai, Junhao Fang, Jiameng Huang, Yueyang Wang, Jinkai Liu,
Yuxuan Zhang

\vspace{0.25em}

\textbf{Corresponding Authors} \\
Jian Yang ,
Zhoujun Li,
Shen Yan,
Wenhao Huang ,
Ge Zhang

\bibliographystyle{plainnat}
\bibliography{main}

\clearpage

\beginappendix

\section{Game Prompt Templates}
\label{app:game-prompts}

This appendix documents the canonical prompts used for the seven games in the
main experiments. The templates are transcribed from
\texttt{test/S3GYM/envs/*/prompt.py}. We normalize whitespace and replace
episode-specific values with angle-bracket placeholders (e.g.,
\texttt{<BOARD>}); the game rules, action spaces, coordinate conventions, and
required answer syntax follow the implementation. At inference time, the
game-specific system prompt is concatenated with the current step prompt and,
for memory-based agents, the available \emph{Hint Messages}.

\definecolor{promptblue}{RGB}{48,91,154}
\definecolor{promptteal}{RGB}{38,123,116}
\definecolor{promptgold}{RGB}{173,116,28}

\newcommand{\promptfield}[2]{\textbf{#1}\enspace #2\par\vspace{0.5pt}}
\newcommand{\gamepromptcard}[6]{%
\begin{tcolorbox}[
  enhanced,
  breakable,
  colback=#1!3!white,
  colframe=#1!72!black,
  boxrule=0.65pt,
  borderline west={2.2pt}{0pt}{#1!88!black},
  arc=1.1mm,
  left=1.8mm,
  right=1.8mm,
  top=0.7mm,
  bottom=0.7mm,
  title={#2},
  coltitle=white,
  colbacktitle=#1!78!black,
  fonttitle=\bfseries\footnotesize,
  before skip=3pt,
  after skip=4pt
]
\footnotesize
\promptfield{System instruction.}{#3}
\promptfield{Dynamic state.}{#4}
\promptfield{Decision instruction.}{#5}
\begin{tcolorbox}[
  colback=white,
  colframe=#1!32!white,
  boxrule=0.45pt,
  arc=0.7mm,
  left=1.5mm,
  right=1.5mm,
  top=0.6mm,
  bottom=0.6mm,
  before skip=3pt,
  after skip=0pt
]
\footnotesize\ttfamily #6
\end{tcolorbox}
\end{tcolorbox}}

\subsection{Shared Interaction Protocol}

Six games prepend the shared base instruction described below, whereas Trust
Evolution uses its own game-specific system instruction. Both memory conditions
retain the same game rules, dynamic state, legal action space, and self-judging
contract. They differ only in how experience from previous exploration episodes
is represented before the current state.

\begin{tcolorbox}[
  enhanced,
  breakable,
  colback=promptblue!3!white,
  colframe=promptblue!72!black,
  boxrule=0.65pt,
  borderline west={2.2pt}{0pt}{promptblue!88!black},
  arc=1mm,
  title={Direct memory: raw-history in-context learning},
  colbacktitle=promptblue!78!black,
  coltitle=white,
  fonttitle=\bfseries\small
]
\small
Direct memory serializes every retained step from previous episodes as an
environment--agent pair,
\texttt{Env: <STATE>} followed by
\texttt{LLM: Answer: <ACTION>, Score: <SELF-SCORE>}.
At action time, the model receives the following composition:
\begin{tcolorbox}[
  colback=white,
  colframe=promptblue!30!white,
  boxrule=0.4pt,
  arc=0.7mm,
  left=1.5mm,
  right=1.5mm,
  top=1mm,
  bottom=1mm,
  before skip=3pt,
  after skip=3pt
]
{\ttfamily\footnotesize
Hint Messages:\par
<RAW HISTORY FROM PREVIOUS EPISODES>\par
\medskip
Game history of current episode:\par
<RAW HISTORY FROM EARLIER STEPS IN THIS EPISODE>\par
\medskip
<GAME SYSTEM PROMPT + CURRENT STEP PROMPT>}
\end{tcolorbox}
The implementation preserves the current-episode history and current game
prompt in full, while truncating the tail of the previous-episode history when
the context budget is exceeded.
\end{tcolorbox}

\begin{tcolorbox}[
  enhanced,
  breakable,
  colback=promptteal!3!white,
  colframe=promptteal!72!black,
  boxrule=0.65pt,
  borderline west={2.2pt}{0pt}{promptteal!88!black},
  arc=1mm,
  title={Summary memory: score-conditioned strategy compression},
  colbacktitle=promptteal!78!black,
  coltitle=white,
  fonttitle=\bfseries\small
]
\small
At step 0 of each episode, Summary memory first makes a separate compression
call over the game rule and truncated previous-episode history:
\begin{tcolorbox}[
  colback=white,
  colframe=promptteal!30!white,
  boxrule=0.4pt,
  arc=0.7mm,
  left=1.5mm,
  right=1.5mm,
  top=1mm,
  bottom=1mm,
  before skip=3pt,
  after skip=3pt
]
{\ttfamily\footnotesize
Rule:\par
<GAME SYSTEM PROMPT>\par
\medskip
History:\par
<TRUNCATED RAW HISTORY FROM PREVIOUS EPISODES>\par
\medskip
Please summarize practical experience and recommendations \ldots\par
Tips: <BULLET-POINTED STRATEGY SUMMARY>}
\end{tcolorbox}
The extracted \texttt{Tips} are cached for the session and reused at later
steps. The action-time input is then
\begin{tcolorbox}[
  colback=white,
  colframe=promptteal!30!white,
  boxrule=0.4pt,
  arc=0.7mm,
  left=1.5mm,
  right=1.5mm,
  top=1mm,
  bottom=1mm,
  before skip=3pt,
  after skip=3pt
]
{\ttfamily\footnotesize
Hint Messages:\par
<SUMMARIZED TIPS>\par
\medskip
<GAME SYSTEM PROMPT>\par
\medskip
Game history of current episode:\par
<RAW HISTORY FROM EARLIER STEPS IN THIS EPISODE>\par
\medskip
<GAME SYSTEM PROMPT + CURRENT STEP PROMPT>}
\end{tcolorbox}
Thus, only cross-episode experience is compressed; the current episode remains
available as raw state--action--score history. In the released implementation,
the game system prompt appears once with the cached summary prefix and again
with the current step prompt.
\end{tcolorbox}

Regardless of memory condition, the model must report both an action and its
self-judged immediate score. The shared final-line contract is

\begin{tcolorbox}[
  enhanced,
  colback=promptgold!4!white,
  colframe=promptgold!72!black,
  boxrule=0.6pt,
  arc=1mm,
  title={Shared response contract},
  colbacktitle=promptgold!78!black,
  coltitle=white,
  fonttitle=\bfseries\small
]
\centering
\ttfamily\small Answer: <ACTION>, Score: <SELF-JUDGED SCORE>.\par
\vspace{2pt}
\rmfamily\footnotesize If the agent determines that the episode has terminated,
it returns \texttt{Answer: END, Score: 0.}
\end{tcolorbox}

The self-judged score is intended to be the reward obtained by the proposed
action under the stated game rules, rather than a free-form confidence value.
Table~\ref{tab:prompt-overview} summarizes the observation and action interfaces
before the full templates are presented.

\begin{table}[h]
\centering
\small
\begin{tabular}{p{0.18\textwidth}p{0.42\textwidth}p{0.31\textwidth}}
\toprule
Game & Main dynamic observation & Valid action form \\
\midrule
Chess & $N\times N$ board with labeled pieces and empty cells & Coordinates for one or more labeled pieces \\
Minesweeper & Partially revealed grid with numbers, flags, and unknown cells & \texttt{uncover}, \texttt{flag}, or \texttt{unflag} \\
Nullify & Indexed list of signed values and arithmetic operators & Two unit indices \\
Tetris & Fixed-cell board and next-block rendering & Drop column and rotation angle \\
Snake & Grid containing walls, head, body, food, and obstacles & \texttt{LEFT/RIGHT/UP/DOWN} \\
Plants-vs-Zombies & Battlefield, sun budget, plants, zombies, and turn state & Zero or more plant placements \\
Trust Evolution & Interaction round and action/payoff history & \texttt{collaborate} or \texttt{cheat} \\
\bottomrule
\end{tabular}
\caption{Observation and action interfaces exposed by the canonical game prompts.}
\label{tab:prompt-overview}
\end{table}

\subsection{State Inference and Deduction}

\gamepromptcard{promptblue}{Chess-style state prediction}{%
You are playing a step-by-step piece-position prediction game on an
$N\times N$ board. Empty cells are denoted by \texttt{.}, pieces are denoted by
letters, coordinates are zero-based, and every piece follows its own hidden
movement rule. Predict the next position of as many pieces as possible from the
observed board-state sequence.}{%
\texttt{Current Game State:} followed by \texttt{<BOARD>}.}{%
Predict the next-step location of each recoverable piece. Use the syntax
\texttt{A:(x,y), b:(x,y), c:(x,y), ...}. The agent may return
\texttt{END} when it elects to stop.}{%
Answer: A:(<ROW>,<COL>), b:(<ROW>,<COL>), ...,
Score: <SELF-JUDGED SCORE>.}

\gamepromptcard{promptblue}{Minesweeper}{%
You are playing Minesweeper on a
\texttt{<BOARD\_ROWS>} $\times$ \texttt{<BOARD\_COLS>} board with
\texttt{<MINES>} hidden mines. Unknown cells are \texttt{?}, flags are
\texttt{F}, revealed counts are \texttt{0--8}, and revealed mines are
\texttt{X}. Coordinates are zero-based. Win by uncovering every safe cell or
correctly flagging every mine; the episode ends at \texttt{<MAX\_EPOCHS>} or
when a mine is uncovered, depending on the evaluation mode.}{%
\texttt{Current Game State:} followed by the partially observed
\texttt{<BOARD>}.}{%
Choose exactly one of \texttt{uncover (r,c)}, \texttt{flag (r,c)}, or
\texttt{unflag (r,c)}. The score is the ratio of correctly flagged mines to
the total number of mines.}{%
Answer: uncover (<ROW>,<COL>), Score: <SELF-JUDGED SCORE>.}

\subsection{Symbolic and Spatial Planning}

\gamepromptcard{promptteal}{Nullify}{%
The board contains indexed independent units. Signed numbers
(\texttt{+n} or \texttt{-n}) can be combined with another unit; available
operators include multiplication, division, square root, square, reciprocal,
floor, and ceiling. Combining equal-magnitude values with opposite signs
produces zero and removes both units. The objective is to eliminate all units,
yielding a final value of zero.}{%
\texttt{Current game board:} followed by one \texttt{index,content} pair per
unit in \texttt{<INDEXED UNITS>}. Unit indices start at zero and are updated
after each combination.}{%
Select the two indices to combine on the current turn. A successful complete
cancellation receives score 1; otherwise the terminal score is 0.}{%
Answer: <INDEX I> <INDEX J>, Score: <SELF-JUDGED SCORE>.}

\gamepromptcard{promptteal}{Tetris}{%
You are playing simplified Tetris on a
\texttt{<BOARD\_WIDTH>} $\times$ \texttt{<BOARD\_HEIGHT>} board. The board
shows fixed blocks as \texttt{*} and empty cells as \texttt{.}; the next-block
panel uses \texttt{\#}. A placed block falls until obstructed, each completed
row is cleared for one point, and the episode terminates when a falling block
extends above the board.}{%
\texttt{Current Game State: <BOARD>} and
\texttt{Next Block: <NEXT\_BLOCK>}.}{%
Choose the one-based drop column of the block's leftmost cell and a clockwise
rotation in \{0, 90, 180, 270\}. The objective is to maximize cleared rows.}{%
Answer: <DROP COLUMN> <ROTATION ANGLE>, Score: <SELF-JUDGED SCORE>.}

\subsection{Reactive Control and Strategic Interaction}

\gamepromptcard{promptgold}{Snake}{%
You control a snake on a
\texttt{<GRID\_WIDTH>} $\times$ \texttt{<GRID\_HEIGHT>} grid. Walls are
\texttt{\#}, the head is \texttt{H}, body cells are \texttt{S}, apples are
\texttt{A}, and obstacles are \texttt{X}. Eating an apple increases both
length and score by one. A wall, self, obstacle, or invalid-action collision
terminates the game under strict rules. The snake cannot reverse direction
and the episode is capped at \texttt{<MAX\_STEPS>}.}{%
\texttt{Current Game State:} followed by \texttt{<BOARD>} and the current
movement direction.}{%
Choose exactly one next movement from \texttt{LEFT}, \texttt{RIGHT},
\texttt{UP}, or \texttt{DOWN}.}{%
Answer: <LEFT|RIGHT|UP|DOWN>, Score: <SELF-JUDGED SCORE>.}

\gamepromptcard{promptgold}{Plants-vs-Zombies}{%
You are playing simplified Plants-vs-Zombies on a
\texttt{<ROWS>} $\times$ \texttt{<COLS>} grid. Each turn gains at least 25
sun; zombies spawn every five turns and move left, becoming stronger over
time. Available plants are Sunflower (\texttt{X}), Peashooter (\texttt{W}),
Three-Line Shooter (\texttt{S}), Wall-nut (\texttt{J}), Torch Stump
(\texttt{H}), and Fire Chili (\texttt{F}), each with the costs and effects
specified in the system prompt. The score increases by one per survived turn;
the game ends when a zombie reaches column $-1$ or after
\texttt{<MAX\_STEPS>}.}{%
\texttt{Current Game State:} followed by \texttt{<BATTLEFIELD>}, including
the turn, sun budget, plant layout, and zombie states.}{%
Place zero or more plants for the next turn. Each placement is
\texttt{PlantType Row Col}; multiple placements are separated by semicolons.
Rows and columns are zero-based, and an empty action is allowed.}{%
Answer: X <ROW> <COL>; W <ROW> <COL>, Score: <SELF-JUDGED SCORE>.}

\gamepromptcard{promptgold}{Trust Evolution}{%
You play a fixed-length repeated interaction against an opponent with a stable
within-episode strategy. The two actions are \texttt{collaborate} and
\texttt{cheat}. Payoffs to the agent are $+2$ for mutual collaboration,
$-1$ for collaborating against cheating, $+3$ for cheating against
collaboration, and $0$ for mutual cheating.}{%
\texttt{Current Game State:} followed by \texttt{<ROUND AND INTERACTION
HISTORY>} for an episode of \texttt{<MAX\_STEPS>} rounds.}{%
Infer the opponent's behavior from the observed interaction and select exactly
one action for the next round.}{%
Answer: <collaborate|cheat>, Score: <SELF-JUDGED SCORE>.}

\end{document}